\documentclass[10pt,twocolumn,letterpaper]{article}

\usepackage{cvis}              

\usepackage{placeins}

\usepackage{float}

\definecolor{cvisblue}{rgb}{0.21,0.49,0.74}
\usepackage[pagebackref,breaklinks,colorlinks,allcolors=cvisblue]{hyperref}
\def\confName{CVIS}
\def\confYear{2026}

\title{From Regression to Classification: Exploring the Benefits of Categorical Representations of Energy in MLIPs}

\author{Ahmad Ali\\
Toronto Metropolitan University\\
Toronto, Ontario, Canada\\
{\tt\small a14ali@torontomu.ca }
}

\begin{document}
\maketitle
\begin{abstract}
Density Functional Theory (DFT) is a widely used computational method for estimating the energy and behavior of molecules. Machine Learning Interatomic Potentials (MLIPs) are models trained to approximate DFT-level energies and forces at dramatically lower computational cost. Many modern MLIPs rely on a scalar regression formulation; given information about a molecule, they predict a single energy value and corresponding forces while minimizing absolute error with DFT's calculations. In this work, we explore a multi-class classification formulation that predicts a categorical distribution over energy/force values, providing richer supervision through multiple targets. Most importantly, this approach offers a principled way to quantify model uncertainty.

In particular, our method predicts a histogram of the energy/force distribution, converts scalar targets into histograms, and trains the model using cross-entropy loss. Our results demonstrate that this categorical formulation can achieve absolute error performance comparable to regression baselines. Furthermore, this representation enables the quantification of epistemic uncertainty through the entropy of the predicted distribution, offering a measure of model confidence absent in scalar regression approaches.
\end{abstract}    
\section{Introduction}
\label{sec:intro}
Density Functional Theory (DFT) is a quantum-mechanical method used in many scientific problems \cite{mlip}. 
By modeling the atomic interactions within a molecule from first principles, it estimates the electronic structure, energy, 
and forces acting on its atoms. DFT calculations are highly precise and are often used as ground-truth labels for many atomic tasks. 
However, DFT is computationally expensive, taking hours or even months to complete. Machine-learning models offer a potential 
alternative that can significantly reduce computational costs. These Machine Learning Interatomic Potentials (MLIPs) are trained 
to approximate DFT's estimates.

MLIPs take as input information about a molecule—such as the positions and types of its atoms—and predict the total energy of the system. 
From this energy, other important attributes, such as atomic forces, can be derived. By using MLIPs to approximate DFT, one can estimate 
key molecular properties much more efficiently \cite{mlip}.

Most MLIPs are formulated as a scalar regression problem: they predict scalar values for energy and forces while minimizing error with 
respect to the ground-truth values. Many state-of-the-art MLIPs are large neural networks with hundreds of millions to billions of 
parameters \cite{uma}. While each layer processes thousands of features, the final output of a particular task is a single scalar, 
which may not be sufficiently informative. This formulation limits both the expressiveness of the training signal and the model’s 
ability to characterize its own uncertainty.

A multi-class classification formulation addresses these limitations by predicting a distribution over discretized energy or force 
values, providing richer supervision and enabling principled uncertainty estimation through entropy. In particular, we use the 
HLGauss method \cite{hlgauss} to convert scalar targets into histogram distributions. These histograms are treated as categorical 
representations and are trained using a multi-class classification objective (cross-entropy loss).

This multi-class formulation provides richer supervision because histogram representations distribute probability mass across 
multiple nearby bins, mitigating overfitting to specific scalar estimates. In addition, the histogram representation leverages 
the inherent ordinal structure of the regression problem, allowing models to smoothly generalize across a continuum of target values. 
This categorical view naturally enables principled uncertainty estimation—an essential capability for scientific modeling—since 
measures such as entropy directly reflect the model’s confidence.

Our cross-entropy formulation achieves performance comparable to the regression baseline, even when evaluated under hyperparameters 
tuned for regression. Although the accuracy is slightly lower, the model produces an uncertainty measure (distribution entropy) that 
correlates with prediction confidence. We additionally perform ablation studies to investigate the design choices behind our 
histogram-based representation.

\section{Methodology}
\label{sec:methodology}

In this section, we introduce MLIPs and describe our histogram representations.  
\subsection{Background.}

MLIP models predict the energy and forces of atoms within a given molecule. The input is a graph representing a molecule, where each node represents an atom.  For the purpose of this paper, we focus on predicting energy only, but a similar method could be used for predicting atomic forces.

The MLIP outputs the energy for each atom $\hat{e}$. DFT computes the energy of a molecule. The ground truth energy for the atom $e$ is computed by dividing (DFT's estimated) molecule energy by the number of atoms in the molecule.

In many MLIPs the loss is chosen to be the absolute error between \cite{uma}:

$$L_{MAE}=|e - \hat{e}|$$

\subsection{Motivation for Classification Formulation}

Classification losses and categorical representations are commonly used in many scalable foundation models such as Large Language Models \cite{gpt}, Vision Transformers \cite{vitpaper}, and Visual Representation Learning \cite{caron2021emerging}. Several reinforcement learning works \cite{stopregressing, dreamerv3} have shown improvements when switching models from continuous squared-error training to categorical representations with classification loss. 

In this paper, we ask if a multi-class classification formulation could potentially offer advantages for MLIPs? 
There are several reasons to believe such an approach can offer advantages for MLIPs.

First and foremost, categorical representations inherently support uncertainty estimation. By predicting a distribution instead of a scalar, the model can quantify its confidence through measures such as entropy, enabling principled assessments of epistemic uncertainty that are not available in standard scalar regression approaches. 

Secondly, converting continuous targets into categorical representations provides richer supervision than predicting a single scalar value. By predicting multiple values, overfitting can be reduced because it encourages the model to distribute probability mass across nearby bins rather than collapsing onto a single point. Furthermore, using the histogram formulation in our paper, we can naturally exploit the ordinal structure of energy values, allowing the model to generalize more smoothly across a range of plausible targets. In addition, predicting multiple values can improve robustness to noise introduced by stochasticity in the underlying data or simulations.

\begin{figure}[!htbp]
    \centering
    \includegraphics[width=1\linewidth]{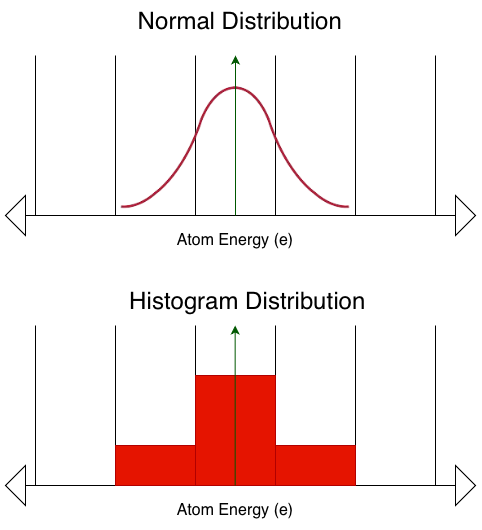}
    \caption{Visualizing scalar target as normal distribution, followed by discretizations of normal distribution into a histogram distribution}
    \label{fig:placeholder}
\end{figure}

\subsection{Histogram Prediction and Loss.}


We now describe a procedure to represent scalar energies as histogram distributions. By virtue of this conversion, the MLIP now outputs a categorical vector representing the energy distribution. Crucially, the MLIP is now trained with cross-entropy loss to match the ground truth histogram distribution \cite{hlgauss}.

Our design closely follows \textit{Histogram Loss with Gaussian smoothing} \cite{hlgauss}. This loss is carefully designed to preserve ordinal relationships between bins.

We denote the histogram as having $K$ bins each of size $w$. Let $c_i$ be the center and $p_i$ be the probability mass in the $i-$th bin. The bin width and centers are chosen so that the histogram covers a specified energy range.

The MLIP outputs a K-dimensional histogram with each bin $i$ having weight $\hat{p}_i$. This is done by taking a softmax of the raw MLIP output logits.

The procedure to convert ground truth (target) scalar energy $e$ to histograms, which involves discretizing a gaussian around the true energy. 

We first create a normal distribution with mean $e$ with fixed variance $\sigma^2$. Then, we discretize this normal distribution into a histogram. In particular, the target probability mass $p_i$ for the $i$-th bin with boundaries $[l_i,\, l_i + w]$ is calculated by integrating the normal distribution over the $i$-th bin interval:

\begin{equation}
p_i = \int_{l_i}^{\,l_i + w}
\frac{1}{\sigma \sqrt{2\pi}}
\exp\!\left( -\frac{(t - y)^2}{2\sigma^2} \right)
\, dt
\label{eq1}
\end{equation}

The MLIP is trained to minimize the cross-entropy divergence between this target
distribution $p$ and the predicted distribution $\hat{p}_i$:

\begin{equation}
\mathcal{L}_{\mathrm{HL}} = - \sum_{i=1}^{K} p_i \log \hat{p}_i(x)
\label{eq2}
\end{equation}

The scalar energy prediction ${e}$ can be recovered from the histogram distribution. This is done by taking expectation of the distribution, which corresponds to taking the weighted sum of the histogram bin centers $c_i$:

\begin{equation}
\hat{e} = \sum_{i=1}^{K} \hat{p}_i(x)\, c_i
\label{eq3}
\end{equation}

\section{Results}

We evaluate discrete representations on the UMA \cite{uma} transformer model and OMoL datasets \cite{omol}. We find that the classification formulation yields comparable or slightly worse performance but provides an accurate estimate of its uncertainty.  

\label{sec:results}

\begin{table*}[t]
\centering
\caption{Evaluation results on OMol25 evaluation splits. Metrics are Energy MAE (meV/atom) and Force MAE (meV/\AA). Best results are in \textbf{bold}.}
\label{tab:main_results}
\resizebox{\textwidth}{!}{%
\begin{tabular}{l cc cc cc cc cc}
\toprule
\textbf{Model} &
\multicolumn{2}{c}{\textbf{Biomolecules}} &
\multicolumn{2}{c}{\textbf{Electrolytes}} &
\multicolumn{2}{c}{\textbf{Metal Complexes}} &
\multicolumn{2}{c}{\textbf{Neutral Organics}} &
\multicolumn{2}{c}{\textbf{Overall}} \\
\cmidrule(lr){2-3} \cmidrule(lr){4-5} \cmidrule(lr){6-7} \cmidrule(lr){8-9} \cmidrule(lr){10-11}
& E & F & E & F & E & F & E & F & E & F \\
\midrule
\textbf{Baseline (MAE)} & \textbf{0.0039} & \textbf{0.0182} & \textbf{0.0080} & \textbf{0.0247} & \textbf{0.0099} & \textbf{0.0623} & 0.0247 & \textbf{0.0352} & \textbf{0.0091} & \textbf{0.0250} \\
\textbf{HL-Gauss (256 bins)} & 0.0065 & 0.0226 & 0.0138 & 0.0309 & 0.0152 & 0.0739 & 0.0147 & 0.0447 & 0.0153 & 0.0310 \\
\textbf{HL-Gauss (128 bins, $\sigma = 0.25$)} & 0.0175 & 0.0223 & 0.0277 & 0.0318 & 0.0248 & 0.0728 & 0.0280 & 0.0453 & 0.0258 & 0.0309 \\
\textbf{HL-Gauss (128 bins, $\sigma = 0.75$)} & 0.0055 & 0.0205 & 0.0110 & 0.0283 & 0.0114 & 0.0679 & \textbf{0.0112} & 0.0400 & 0.0122 & 0.0284 \\
\textbf{HL-Gauss (128 bins, $\sigma = 2.00$)} & 0.0089 & 0.0264 & 0.0196 & 0.0361 & 0.0229 & 0.0815 & 0.0219 & 0.0544 & 0.0188 & 0.0361 \\
\bottomrule
\end{tabular}%
}
\end{table*}
\subsection{Experimental Setup.}

All models utilize the Universal Model for Atoms (UMA) architecture \cite{uma}. The DFT dataset is taken from the Open Molecules 2025 (OMol25) dataset \cite{omol}. We compare a baseline model trained with the standard scalar absolute error regression objective against models equipped with our Histogram Loss (HL-Gauss) classification head. We use a softmax temperature of 2.0.

All models use the same pre-training hyperparameters (AdamW, Cosine scheduler) and are evaluated on the official OMol25 validation set across biomolecules, electrolytes, metal complexes, and neutral organics. All models have 150M parameters, corresponding to the UMA-S variant.


We train the model on the basic \textit{Direct Force Pre-training Stage} of the UMA \cite{uma}. In this stage, forces are also predicted in addition to the energy. The forces are predicted by a separate head. The final training objective is a weighted sum of the cross-entropy loss for energy:($\mathcal{L}_{\mathrm{CE}}$) and the regression loss for forces ($\mathcal{L}_{\mathrm{Force}}$):

\begin{equation}
\mathcal{L}_{\mathrm{total}} = 0.7\,\mathcal{L}_{\mathrm{CE}} + 0.3\,\mathcal{L}_{\mathrm{Force}}
\label{eq4}
\end{equation}

\subsection{Main Results.}
We report the Mean Absolute Error (MAE) for energy and forces across the standard evaluation splits. Table \ref{tab:main_results} summarizes the performance of the baseline regression model versus the HL-Gauss classification models.

We find that the classification approach, particularly the 128-bin configuration, achieves comparable accuracy ($\pm 0.0015$) with the scalar baseline. With more hyperparameter tuning, it's possible that we get better performance with HL-Gauss loss. However, the baseline outperforms HL-Gauss on all the tasks except for one.

\subsection{Ablations.}

Our experiments focus on an ablation study to determine the impact of bin resolution on accuracy. 

We investigate the sensitivity of the classification head to two key hyperparameters: bin resolution and the Gaussian smoothing factor ($\sigma$).
\begin{itemize}
    \item \textbf{Bin Resolution:} We compare bin counts of 128 and 256. The 128 bin ablation slightly outperforms the 256 bin ablation. 
    
    \item \textbf{Target Distribution ($\sigma$):} Using the 128-bin model, we ablate $\sigma$ multipliers of $\{0.25, 0.75, 2.0\}$. We find that $\sigma = 0.75 \times \text{bin\_width}$ performs best. A small sigma ($0.25$) creates "spiky" targets approaching a one-hot distribution. Whereas, a large sigma ($2.0$) over-smooths the target. These findings match \cite{stopregressing}, which recommends a $\sigma$ multiplier of $0.75$ .

\end{itemize}

\subsection{Estimating Uncertainty.}
An advantage of the HL-Gauss head is the ability to quantify epistemic uncertainty via the entropy of the predicted distribution $e$. Figure 2 shows the Pearson Correlation Coefficient between the model's prediction entropy and the absolute error of its energy prediction on the training set. We observe that the uncertainty is positively correlated with error, indicating that the model metric accurately captures the model's epistemic uncertainty. However, the correlation is quite noisy, indicating that entropy is an imprecise measure of uncertainty. An interesting experiment would be to evaluate correlation on larger datasets or later on in training.  

\begin{figure}[h!]
    \centering
    \includegraphics[width=\linewidth]{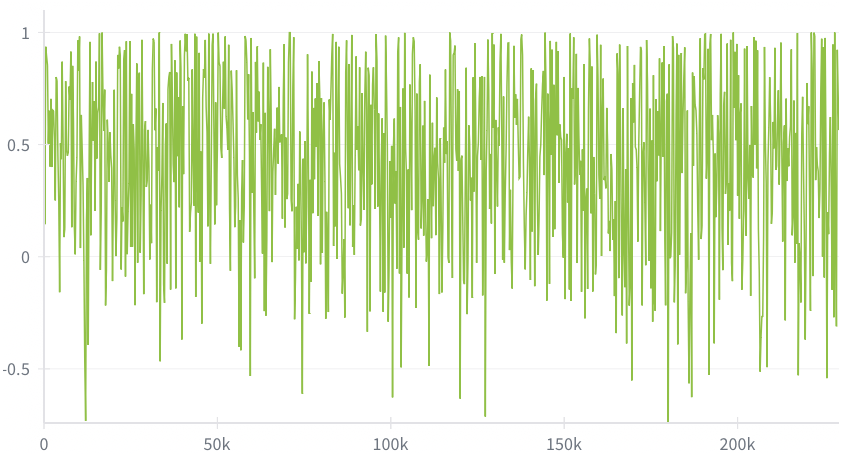}
    \caption{Pearson Correlation Coefficient between Histogram Entropy and the absolute error across training batches. The x-axis is training step, y-axis is the correlation coefficient.}
    \label{fig:entropy_corr}
\end{figure}

\section{Conclusion}

In this work, we investigated the efficacy of replacing standard scalar regression with a categorical energy classification head for training Machine Learning Interatomic Potentials (MLIPs). By predicting a histogram of energy distributions using the Universal Model for Atoms (UMA) architecture, we demonstrated that a cross-entropy formulation—specifically the Histogram Loss with Gaussian smoothing (HL-Gauss)—achieves performance comparable to traditional regression baselines.

Beyond raw accuracy, this distributional approach provides the novel advantage of quantifying epistemic uncertainty via prediction entropy. However, we observe that the correlation between this entropy and absolute prediction error is noisy, suggesting that while the method yields a confidence signal, it does not yet reliably predict specific model errors. Future work should focus on refining this uncertainty measure and exploring if the stability benefits of cross-entropy enable better scaling for larger foundation models.





{
    \small

}

\end{document}